\documentclass[
]{ceurart}

\usepackage{listings}
\begin{document}

\copyrightyear{2024}
\copyrightclause{Copyright for this paper by its authors.
  Use permitted under Creative Commons License Attribution 4.0
  International (CC BY 4.0).}

\conference{Version as accepted at the BioASQ Lab at CLEF 2024}

\title{Can Open-Source LLMs Compete with Commercial Models? Exploring the Few-Shot Performance of Current GPT Models in Biomedical Tasks}

\title[mode=sub]{Notebook for the BioASQ Lab at CLEF 2024}

\author[1]{Samy Ateia}[%
email=Samy.Ateia@stud.uni-regensburg.de,
orcid=0009-0000-2622-9194,
url=https://www.uni-regensburg.de/language-literature-culture/information-science/team/samy-ateia-msc,
]
\author[1]{Udo Kruschwitz}[%
email=udo.kruschwitz@ur.de,
url=https://www.uni-regensburg.de/language-literature-culture/information-science/team/udo-kruschwitz/,
orcid=0000-0002-5503-0341,
]

\address[1]{Information Science, University of Regensburg, Universitätsstraße 31, 93053, Regensburg, Germany}

\begin{abstract}
Commercial large language models (LLMs), like OpenAI's GPT-4 powering ChatGPT and Anthropic's Claude 3 Opus, have dominated natural language processing (NLP) benchmarks across different domains. New competing Open-Source alternatives like Mixtral 8x7B or Llama 3 have emerged and seem to be closing the gap while often offering higher throughput and being less costly to use. Open-Source LLMs can also be self-hosted, which makes them interesting for enterprise and clinical use cases where sensitive data should not be processed by third parties. We participated in the 12th BioASQ challenge, which is a retrieval augmented generation (RAG) setting, and explored the performance of current GPT models Claude 3 Opus, GPT-3.5-turbo and Mixtral 8x7b with in-context learning (zero-shot, few-shot) and QLoRa fine-tuning. We also explored how additional relevant knowledge from Wikipedia added to the context-window of the LLM might improve their performance. Mixtral 8x7b was competitive in the 10-shot setting, both with and without fine-tuning, but failed to produce usable results in the zero-shot setting. QLoRa fine-tuning and Wikipedia context did not lead to measurable performance gains. Our results indicate that the performance gap between commercial and open-source models in RAG setups exists mainly in the zero-shot setting and can be closed by simply collecting few-shot examples for domain-specific use cases. The code needed to rerun these experiments is available through GitHub*.
\end{abstract}

\tnotetext[1]{\url{https://github.com/SamyAteia/bioasq2024}}


\maketitle

\section{Introduction}
Over the course of 2023, NLP benchmarks in various domains were dominated by commercial LLMs that are only accessible via APIs that make it difficult to do transparent and reproducible research \cite{chen2024chatgpts}. They also might not be usable in clinical or enterprise use cases where sensitive data cannot be shared with third parties. In March 2023, OpenAI had to briefly take ChatGPT offline because they were accidentally leaking user-messages\footnote{\url{https://web.archive.org/web/20240503032019/https://openai.com/index/march-20-chatgpt-outage/}}. In April 2023 Samsung had to ban the use of ChatGPT because employees shared sensitive data with the system\footnote{\url{https://web.archive.org/web/20240518030412/https://techcrunch.com/2023/05/02/samsung-bans-use-of-generative-ai-tools-like-chatgpt-after-april-internal-data-leak/}}. These examples show that there are real issues with the confidentiality with these services, while no competitive offline alternatives existed in early 2023.

But some companies like Mistral\footnote{\url{https://mistral.ai/news/mixtral-of-experts/}} and Meta\footnote{\url{https://llama.meta.com/llama3/}} started to publish their state-of-the-art (SOTA) LLMs with permissive licenses and are making the model weights downloadable. This makes these models especially interesting for research directed at enterprise and clinical applications, as they can be hosted on your hardware in a controlled environment.

As enterprise use cases often have to deal with domain-specific data that is not publicly available to the LLMs during pre-training, retrieval augmented generation (RAG) \cite{lewis2020retrieval} is often used to enable these models to understand new concepts and be more helpful and grounded in their responses \cite{balaguer2024rag}. Several software vendors are publishing marketing articles to advertise the usefulness of their solutions to enable RAG for enterprises \footnote{\url{https://cohere.com/blog/five-reasons-enterprises-are-choosing-rag}}\footnote{\url{https://www.pinecone.io/learn/retrieval-augmented-generation/}}\footnote{\url{https://gretel.ai/blog/what-is-retrieval-augmented-generation}}.

The BioASQ challenge is a great example of a RAG setup in a specialized domain, as the participating systems first have to find relevant biomedical papers from PubMed and extract snippets that are later used to generate answers for biomedical questions.

We set out to explore the usefulness and competitiveness of open-source models to the current SOTA commercial offerings in a typical domain-specific RAG setup represented by the BioASQ challenge. Compared to our last year's approach where we only looked at the zero-shot performance of commercial models, we now explored few-shot learning because we saw that this enables open-source models to better follow instructions while also improving overall performance. 

Another aspect that we explored was how additional relevant context retrieved from Wikipedia might aid the models in generating useful answers or relevant queries, as they might be limited in their biomedical knowledge about entities and their synonyms.

\subsection{BioASQ Challenge}
BioASQ is "a competition on large-scale biomedical semantic indexing and question answering"\cite{BioASQ2024overview}. It is held as a lab at the Conference and Labs of the Evaluation Forum (CLEF) conference\footnote{\url{https://clef2024.clef-initiative.eu/}}. The current 2024 workshop is the 12th installment of the BioASQ competition\footnote{\url{http://www.bioasq.org/}}. 

The 12th BioASQ challenge comprises several tasks:
\begin{itemize}
    \item BioASQ Task Synergy On Biomedical Semantic QA For Developing Issues \cite{task12bSynergy2024overview}
    \item BioASQ Task B On Biomedical Semantic QA \cite{task12bSynergy2024overview} 
    \item ioASQ Task MultiCardioNER On Mutiple Clinical Entity Detection In Multilingual Medical Content \cite{multicardioner2024overview}
    \item BioASQ Task BioNNE On Nested NER In Russian And English \cite{bionne2024overview}
\end{itemize}

We participated in \emph{Task B} and \emph{Synergy}\cite{task12bSynergy2024overview}. For \emph{Task B} the participants' systems receive a list of biomedical questions that should be answered with a short paragraph style answer and some require an additional exact answer which can be one of 3 formats, yes/no, factoid (a list of up to 5 entities) or list (a list of up to 200 entities). Additionally, the systems first have to retrieve relevant papers from the PubMed annual baseline and extract relevant snippets from these papers that could aid in answering the questions. This retrieval subtask of \emph{Task B} is called \emph{Phase A} while the actual question answering subtask is called \emph{Phase B}. For \emph{Phase B} the systems also receive a set of gold snippets and documents that should help them answer the question. \emph{Task B} was scheduled in 4 batches with two weeks in between and ran from March 28 to May 11.

In the 12th installment, another \emph{Phase A+} was introduced, where the systems were supposed to provide answers to the questions before the gold snippets and documents were provided, relying solely on their own retrieved documents and snippets.

For the \emph{Synergy} task, the systems receive a similar list of questions, for which they also have to retrieve useful papers and extract snippets and as soon as a question is marked as ready to answer, they also need to submit answers in the same format as for \emph{Task B}. The difference between \emph{Synergy} and \emph{Task B} is that initially in the first round no gold set of documents and snippets is provided, instead the submitted documents and snippets by the systems are evaluated by biomedical experts and selected as gold reference items for subsequent rounds. This also means that the same questions might be reintroduced in subsequent rounds, possibly with additional questions and positive and negative feedback on the previously submitted documents.

Following this introduction, we will highlight some related work in Section \ref{Related}, describe our methodology in Section \ref{Methods}, report our results in Section \ref{Results} and discuss them in Section \ref{Discussion}. Section \ref{Ethics} will present some ethical considerations, and Section \ref{Conclusions} offers our conclusions. 

\section{Related Work}\label{Related}
We will briefly introduce the related work that led to the creation of the evaluated models, as well as the approaches that inspired our methodology.

\subsection{GPT Models}
Nearly all the popular SOTA LLMs that are used across various NLP tasks and use cases today are based on the transformer architecture \cite{vaswani2017attention}. With the generative pretrained transformer (GPT) \cite{radford2018improving} being a popular variant. These models undergo pre-training on vast amounts of text by solving the next-token prediction task \cite{radford2018improving}. Afterwards, the models are fine-tuned to align with human preference data \cite{ouyang2022training} which enables them to follow instructions and be useful in direct interactions with users.

OpenAI was the first company that released such a fine-tuned model to the public in November 2022\footnote{\url{https://web.archive.org/web/20240502090536/https://openai.com/index/chatgpt/}} which sparked massive interest in generative artificial intelligence research and products. Their latest model at the time of writing that is powering their ChatGPT product was GPT-4 \cite{openai2023gpt4}. One interesting competitor model that we also used during this competition is Claude 3 Opus\footnote{\url{https://web.archive.org/web/20240516173322/https://www-cdn.anthropic.com/de8ba9b01c9ab7cbabf5c33b80b7bbc618857627/Model_Card_Claude_3.pdf}} by Anthropic, which reached GPT-4 level performance (GPT-4-0125-preview) at the time of the BioASQ competition\footnote{\url{https://chat.lmsys.org/?leaderboard}}. The exact architecture of GPT-4 and Claude Opus 3 and other commercial models is unknown.

GPT-4 is the most expensive and slowest model that OpenAI is offering via their API service. A more affordable alternative that they are offering is GPT-3.5-turbo. We compared both these models' performance in last year's BioASQ competition and were able to show that GPT-3.5-turbo was sometimes performing better than GPT-4 in some question formats and subtasks of the competition \cite{ateiakruschwitz}. 

For this year's BioASQ competition we also used Mixtral 8x7B \cite{jiang2024mixtral} a downloadable open-source model (Apache 2.0 license) which uses a type of Mixture-of-Experts Architecture \cite{jacobs1991adaptive}\cite{eigen2014learning}. This architecture offers higher computational efficiency by routing requests to expert subnetworks to generate a response. Since only some specialized experts are active during the generation, less computation and memory is needed to serve the request.

\subsection{Few and Zero-Shot Learning}
Few-Shot learning is the ability of LLMs to learn how to solve a new problem that they were not specifically fine-tuned for by only showing them a few examples. When GPT-3 was first introduced, its impressive few-shot learning abilities made the concept popular \cite{brown2020language} because it greatly reduces the need for expensive training data.

Zero-shot learning \cite{palatucci2009zero} takes this concept a step further by only requiring an abstract task description or direct question, which leads the model to ideally generate a useful completion that solves the task at hand \cite{liu2023pre}. In last year's BioASQ competition, we were able to win some batches while only using zero-shot learning with SOTA commercial models \cite{ateiakruschwitz}. 

\subsection{Adapter Fine-Tuning}
Current LLMs have billions of parameters and require specialized hardware with enough GPUs and VRAM to hold all the model weights in memory. For example, "16-bit finetuning of a LLaMA 65B parameter model requires more than 780 GB of GPU memory"\cite{dettmers2023qlora}. Since this makes fine-tuning these models prohibitively expensive for many researchers and users, several clever techniques have been invented to reduce these hardware requirements. We wanted to fine-tune Mixtral 8x7B, which roughly takes up the memory of a 47B model\footnote{\url{https://mistral.ai/news/mixtral-of-experts/}}. 

One popular approach is QLora by Dettmers et al. \cite{dettmers2023qlora} where the model weights are quantized to 4 bits and frozen and only some low rank adapters (LoRa)\cite{hu2022lora} are fine-tuned. This would enable fine-tuning Mixtral 8x7B, for example, on only two RTX A6000 GPUs with 2x 48 GB of VRAM. 

\subsection{Retrieval Augmented Generation (RAG)}
Retrieval augmented generation (RAG) is a technique \cite{lewis2020retrieval} that combines information retrieval with language models to enhance their ability to generate relevant and factual text. In RAG, the language model is augmented with an external knowledge base or other information source, such as a collection of documents or web pages. When generating text, the model first retrieves relevant information, based on the input query, and then uses that information to guide the generation process. This process is applied in the BioASQ challenge, where the relevant information source is the annual baseline of PubMed.

RAG has been shown to improve the factual accuracy of generated text compared to standalone language models \cite{shuster2021retrieval}. It allows the model to access a vast amount of external knowledge and incorporate it into the generated output. RAG is particularly useful for tasks that require domain-specific knowledge or up-to-date information \cite{balaguer2024rag}.

\subsection{Professional Search}
Professional search is conducted in a professional context, often to aid in work-related research tasks \cite{Tait2014}. In some professional search settings, highly trained specialists are needed to create documented and reproducible search strategies, this sets professional search apart from everyday web search \cite{DBLP:journals/sigir/VerberneHKWLRV18}. The BioASQ challenge exemplifies one possible professional search setting where biomedical experts aim to find answers to domain-specific questions with sufficient evidence. 

Other examples of professional search might be systematic reviews \cite{MACFARLANE2022200091}, patent-search or search conducted by recruitment professionals \cite{doi:10.1177/02663821211034079}. All of these settings might require complex search strategies, where the search expert makes use of a query syntax involving boolean operators on specific search fields. Systematic reviews, for example, also require the search to be explainable and reproducible, which makes it difficult to use advanced vector-based retrieval techniques. Formulating traditional queries but with large language models that might be able to expand synonyms and related terms based on their semantic representations is therefore an interesting approach that might aid in professional search settings. We set out to explore this approach in the BioASQ challenge.


\section{Methodology}\label{Methods}

\subsection{Model}
In this year's BioASQ competition, we looked at the commercial offerings GTP-3.5-turbo and GPT-4 from OpenAI and also used Antrophics Claude 3 Opus, which was at the time of the run submissions the only other  model that was on a level with GPT-4 according to the LMSYS Chatbot Arena Leaderboard \cite{chiang2024chatbot}. 

Since last year's BioASQ competition, some competitive Open-Source models were published. The most notable ones being the Llama series models by Meta \cite{touvron2023llama}, with the latest being Llama 3 \cite{llama3modelcard}. Llama comes with its own custom License which is quite permissive, but even though commercial use is allowed under this license, as long as the monthly user base does not exceed 700 million users, the license might not be straightforward to adopt for enterprise use cases when licenses have to be pre-approved by a legal team. 

When we prepared our runs for the competition, the best-performing model with a permissive open-source license (Apache 2.0) on the LMSYS leaderboard was Mixtral 8x7B \cite{jiang2024mixtral}. The model also has a large context length of 32k tokens, which makes it especially interesting for RAG use cases or few-shot learning. We therefore choose Mixtral 8x7B as our open-source competitor model for this competition. During the competition, the newer Mixtral 8x22B model was also published, and we used it in some batches of Task B.

We used the commercial hosting service fireworks.ai\footnote{\url{https://fireworks.ai/}} to access and fine-tune Mixtral 8x7B, as the provided speed was very high and costs for their API usage were low.

\subsection{Synergy}
We downloaded and indexed the annual PubMed baseline from the official website\footnote{\url{https://pubmed.ncbi.nlm.nih.gov/download/}}. We indexed both the title and the abstract of all papers in separate fields of our index using the built-in English analyzer of Elasticsearch\footnote{\url{https://www.elastic.co/guide/en/elasticsearch/reference/current/analysis-lang-analyzer.html\#english-analyzer}}. For every round of synergy, the most recent snapshots of 2024 up to the date considered in that round were downloaded and indexed in another similar index, which was then also searched during the runs.

For synergy, we used both gpt-4-0125-preview and gpt-3.5-turbo-0125, the newest available versions of OpenAI's GPT-4 and GPT-3.5-turbo at the time of the competition. We used \textbf{2-shot learning} to generate queries for our Elasticsearch PubMed index and \textbf{zero-shot} learning for extracting and reranking snippets as well as answering questions. We also wanted to use Mixtral 8x7B in this task, but the model was unable to follow instructions well enough to produce usable runs, especially in the zero-shot setting.

\begin{lstlisting}[float,language=python,frame=single,caption={Query Expansion Prompt},label=expansionListing,breakindent=0pt,columns=fullflexible]
{"role": "user", "content": f"""Turn the following biomedical question into an effective elasticsearch query using the query_string query type by incorporating synonyms and additional terms that closely relate to the main topic and help reduce ambiguity. Focus on maintaining the query's precision and relevance to the original question, the index contains the fields 'title' and 'abstract', return valid json: '{question}'"""}
\end{lstlisting}

Given a question, we prepended the prompt in Listing \ref{expansionListing} with two examples where the same prompt contained other questions, for example "Is CircRNA produced by back splicing of exon, intron or both, forming exon or intron circRNA?" and an ideal completion in form of an Elasticsearch query for the query\_string endpoint of Elasticsearch for which an example can be seen in Listing \ref{expansionListingCompletion}. We sent both the examples and the prompt with our actual question to the model and received back a JSON object that could be used to directly query our index.
\begin{lstlisting}[float,language=python,frame=single,caption={Query Expansion Competion Example},label=expansionListingCompletion,breakindent=0pt,columns=fullflexible]
 {"role": "assistant", "content": """
    {
        "query": {
            "query_string": {
            "query": "(CircRNA OR \"circular RNA\") \"back splicing\" exon OR intron",
            "fields": [
                "title^10",
                "abstract"
            ],
            "default_operator": "and"
            }
        },
        "size": 50
        }
    """},
\end{lstlisting}

We ran the generated query to retrieve the top 50 relevant documents from Elasticsearch. We filtered out documents that were marked as irrelevant in the feedback file for the synergy round. We then sent each remaining article alongside the question to the model and used a zero-shot prompt to extract a list of relevant snippets from the article. We then used string matching to insure the returned snippets were actually present in the article title or abstract and to calculate the offsets. 

We collected all relevant snippets from all potentially 50 articles and filtered out articles as irrelevant where the model did not extract any snippets from. We then prompted the model to select the top 10 snippets ranked by helpfulness from the set of snippets. We finally reranked the retrieved articles according to the snippet order returned by this step.

In the question answering step, we used the identical zero-shot prompts from our last year's participation in Task B for this year's Synergy task \cite{ateiakruschwitz}. And we also merged the already deemed relevant snippets from the feedback files into our list of snippets that we passed on to the modal alongside the question prompt.

We also sent the same initial system prompt that we used last year \cite{ateiakruschwitz} to the models. For the parameters, we set the \emph{temperature} parameter to 0 to reduce randomness in the completion and supplied a seed parameter which is a new feature offered by OpenAI that can help maximize reproducibility for the model output, but determinism is still not guaranteed\footnote{\url{https://platform.openai.com/docs/api-reference/chat/create\#chat-create-seed}}. We also used the new response\_format parameter to insure the model produced valid JSON responses for the prompts where we needed it\footnote{\url{https://platform.openai.com/docs/api-reference/chat/create\#chat-create-response_format}}. The exact python notebooks with all the implementation details and prompts used are available in our GitHub repository.

\subsection{Task 12 B}
For Task B we reused the indexed PubMed annual snapshot that we created from the synergy task. We also switched the models, we added Mixtral 8x7B Instruct v0.1 as an open-source model, and we used Claude 3 Opus instead of GPT-4 because it became available shortly before the Phase started, and we had access to a free beta evaluation account. 

In batches 1 \& 2 we also explored the fine-tuning service of OpenAI and created 6 fine-tuned versions of GPT-3.5-turbo for each sub problem that our system had to solve. We also created 6 QLoRa fine-tuned versions of Mixtral 8x7B Instruct v0.1 using the fine-tuning service of fireworks.ai. The training sets were created from the supplied training data. We also adjusted our code  to be able to use these training files as sources for few-shot examples.

The sub problems that we sampled training sets for were:
\begin{itemize}
    \item Snippet Extraction
    \item Snippet Reranking \& Selection
    \item Summary Question Answering
    \item Exact yes/no Question Answering
    \item Exact factoid Question Answering
    \item Exact list Question Answering
\end{itemize}

In batches 3\&4 we explored how we could use the models to retrieve relevant additional context from Wikipedia. We hypothesized that supplying these models with knowledge about relevant entities in the questions might improve their ability to generate correct answers. We wanted to explore the approach of retrieving such additional information about entities from a Wiki, because such a wiki like could also be generated in an enterprise setting, potentially closing the knowledge gap for entities that the models didn't encounter during pre-training. We chose Wikipedia as a knowledge base even though the concepts described there might not be novel to the models because it was easy to use, and we hoped that we could observe an effect even with known concepts. We suspected that these models might just "know" which Wikipedia articles might be relevant to a question because they might have been highly trained on Wikipedia and links to Wikipedia articles. The exact zero-shot prompt that was used for finding relevant Wikipedia articles can be seen in Listing \ref{wikiListing}.

\begin{lstlisting}[float,language=python,frame=single,caption={Wikipedia Retrieval Prompt},label=wikiListing,breakindent=0pt,columns=fullflexible]
prompt = f"""
    Given the question "{question}", identify existing Wikipedia articles that offer helpful background information to answer this question. 
    Ensure that the titles listed are of real articles on Wikipedia as of your last training cut-off. Wrap the confirmed article titles in hashtags (e.g., #Article Title#). 
    Provide a step-by-step reasoning for your selections, ensuring relevance to the main components of the question.

    Step 1: Confirm the Existence of Articles
    Before listing any articles, briefly verify their existence by ensuring they are well-known topics generally covered by Wikipedia.

    Step 2: List Relevant Wikipedia Articles
    After confirming, list the articles, wrapping the titles in hashtags and explaining how each article is relevant to the question.
    """
\end{lstlisting}

The titles of the Wikipedia articles returned by this prompt were extracted and, if the Wikipedia articles actually existed, their content was retrieved and concatenated. Finally, the concatenated articles were again sent to the model, and it was prompted to make a concise summary to help answer the question. This summary was then added as additional context to all subsequent prompts such as query generation or snippet extraction, reranking and question answering.

\subsubsection{Phase A}
In Phase A, we changed the way we prompted the models for queries compared to the Synergy task. Instead of expecting the whole valid JSON query object back, we only prompted the models to create the query string in the valid query\_string query syntax to pass on to the Elasticsearch endpoint and manually controlled weighting of the fields and the document return size. 

We then used the questions from batch 1 from last year's task, that were provided in the development set, to create a set of Elasticsearch queries with Claude 3 Opus. Finally, we ran these queries and evaluated the returned documents against the gold set to select 10 queries with the highest f1 score as few-shot examples for all models. We used 10 examples for most of our few-shot tasks because they fit in most models context lengths, with GPT-3.5-turbo having the smallest context length of 16k tokens.

The rest of our approach to retrieving relevant documents and snippets was mostly in line with our approach from the Synergy task, except that we re-added the step that we also used last year when we prompted the models for an improved  query when a generated query did not return any results. 

Compared to our system from last year, we changed the query expansion/generation prompt, added the possibility to prepend few-shot examples to the prompts and used different models where some were also fine-tuned. We also reranked snippets instead of titles, and filtered and ranked the articles according to their extracted snippets. We also used our own Elasticsearch index of the PubMed annual baseline instead of the online PubMed search endpoint.

\subsubsection{Phase A+ and Phase B}

Our approach for Phase A+ and Phase B were mostly identical. We used the same prompts and few-shot examples taken from the sampled training sets for model fine-tuning. The only difference was the relevant snippets provided as context to the models. For Phase A+ we used the same snippets as input for all models, these were taken from the run in Phase A where the most snippets were found. We opted to use the same snippets, as we wanted to be able to compare the performance of the different models in Phase A+ isolated from their performance in Phase A. For Phase B we took the gold snippets provided in the run file as input.

The prompts used for Phase A+ and Phase B were identical to the question answering prompts from the Synergy task and our last year's approach. We only added the option to add additional context before the snippets and supply few-shot examples. In batches 1 \& 2 we compared the performance of fine-tuned models with their non-fine-tuned counterparts, and in batches 3 \& 4 we compared systems with additional relevant context taken from Wikipedia with systems that did not have this context.

\section{Results}\label{Results}

We participated with our systems in two tasks of the BioASQ challenge: \emph{Synergy} and \emph{Task B On Biomedical Semantic QA}. For the synergy task, we report the results only for batch 3 and 4 as we were unable to participate in earlier batches. For Task B we competed in all batches and report the results of sub-tasks A (Retrieval), A+ (Q\&A with own retrieved documents) and B (Q\&A with gold documents). The results presented in this section are only preliminary, as the manual assessment of the system responses
by the BioASQ team of biomedical experts is still ongoing. The final results will be available on the BioASQ homepage once the manual assessment is finished\footnote{\url{http://participants-area.bioasq.org/results/}}.

\subsection{Synergy}
We participated with 2 systems in batch 3 and 4 of the Synergy task. The full result table is accessible on the BioASQ website\footnote{\url{http://participants-area.bioasq.org/results/synergy_v2024/}}. The two systems both used the same 2-shot query expansion and zero-shot Q\&A approach but with different commercial models. The system names and corresponding models are listed below:

\begin{itemize}
    \item UR-IW-2: gpt-4-0125-preview.
    \item UR-IW-3: gpt-3.5-turbo-0125
\end{itemize}

We tried to also use Mixtral 8x7B Instruct v0.1 as an open-source alternative in this task, but the model was unable to follow the zero-shot prompts for query expansion and extracting the snippets consistently enough to produce a submittable run file. Our retrieval approach for query expansion and filtering and reranking via extracted snippets was \textbf{not competitive} in the document retrieval stage of the task, with both models performing similarly poorly compared to the top competitors, as indicated in Table \ref{tab:synergyDocumentsBatches}. The official metrics to rank systems in each subtask are highlighted in bold in the following tables\footnote{"Top Competitor" are the systems that took the first position in a round or batch that are not ours. They are added as a reference point for the reported metrics. When "Top Competitor" is missing in a reported batch, one of our systems was the best-performing one.}.

\begin{table}[h!]
\caption{Task 12 Synergy, Document Retrieval, Batches 3-4 }
\footnotesize
\label{tab:synergyDocumentsBatches}
\begin{tabular}{|c|c|c|c|c|c|c|c|}
\hline
Batch & Position & System & Precision & Recall & F-Measure & \textbf{MAP} & GMAP \\
\hline
Test round 3 & 1 of 9 & Top Competitor & 0.2156 & 0.2568 & 0.1981 & \textbf{ 0.1769 } & 0.0195 \\
 & \textbf{6} of 9 & UR-IW-3 & 0.1076 & 0.0619 & 0.0675 & \textbf{ 0.0532 } & 0.0003 \\
 & \textbf{7} of 9 & UR-IW-2 & 0.1076 & 0.0619 & 0.0675 & \textbf{ 0.0532 } & 0.0003 \\
\hline
Test round 4 & 1 of 13 & Top Competitor & 0.1651 & 0.1671 & 0.1459 & \textbf{ 0.1308 } & 0.0070 \\
 & \textbf{5} of 13 & UR-IW-3 & 0.0912 & 0.0831 & 0.0790 & \textbf{ 0.0664 } & 0.0006 \\
 & \textbf{7} of 13 & UR-IW-2 & 0.0792 & 0.0746 & 0.0628 & \textbf{ 0.0514 } & 0.0003 \\
\hline
\end{tabular}
\end{table}

For snippet extraction, the performance of our approach was also poor, except for batch 4 where gpt-3.5-turbo-0125 was able to achieve second place as can be seen in Table \ref{tab:synergySnippetsBatches}. Gpt-4-0125-preview was unable to extract any snippets in the same batch, which was unusual and could have been due to issues OpenAI had with serving the preview model via their API.

\begin{table}[h!]
\caption{Task 12 Synergy, Snippet Extraction, Batches 3-4}
\footnotesize
\label{tab:synergySnippetsBatches}
\begin{tabular}{|c|c|c|c|c|c|c|c|}
\hline
Batch & Position & System & Precision & Recall & F-Measure & \textbf{MAP} & GMAP \\
\hline
Test round 3 & 1 of 9 & Top Competitor & 0.1751 & 0.1444 & 0.1356 & \textbf{ 0.1811 } & 0.0019 \\
 & \textbf{6} of 9 & UR-IW-3 & 0.0795 & 0.0467 & 0.0474 & \textbf{ 0.0567 } & 0.0001 \\
 & \textbf{7} of 9 & UR-IW-2 & 0.0795 & 0.0467 & 0.0474 & \textbf{ 0.0567 } & 0.0001 \\
\hline
Test round 4 & 1 of 13 & Top Competitor & 0.1241 & 0.1103 & 0.0982 & \textbf{ 0.1003 } & 0.0018 \\
 & \textbf{2} of 13 & UR-IW-3 & 0.0966 & 0.0852 & 0.0741 & \textbf{ 0.0989 } & 0.0005 \\
 & \textbf{13} of 13 & UR-IW-2 & - & - & - & \textbf{ - } & - \\
\hline
\end{tabular}
\end{table}

For the question-answering stage, both models achieved perfect scores in the exact yes/no answer format, see Table \ref{tab:synergyYesNoBatches}. For factoid answers, gpt-4-0125-preview was able to take first place in batch 4 and achieved higher placements than gpt-3.5-turbo-0125 over both batches, as can be seen in Table \ref{tab:synergyFactoidBatches}. A similar difference is also observable in Table \ref{tab:synergyListBatches} for the list answer results.

\begin{table}[h!]
\caption{Task 12 Synergy, Q\&A exact Yes/No, Batches 3-4}
\footnotesize
\label{tab:synergyYesNoBatches}
\begin{tabular}{|c|c|c|c|c|c|c|}
\hline
Batch & Position & System & Accuracy & F1 Yes & F1 No & \textbf{Macro F1} \\
\hline
Test round 3 & \textbf{1} of 9 & \textbf{UR-IW-3} & 1.0000 & 1.0000 & 1.0000 & \textbf{ 1.0000 } \\
 & \textbf{1} of 9 & UR-IW-2 & 1.0000 & 1.0000 & 1.0000 & \textbf{ 1.0000 } \\
\hline
Test round 4 & \textbf{1} of 13 & \textbf{UR-IW-3} & 1.0000 & 1.0000 & 1.0000 & \textbf{ 1.0000 } \\
 & \textbf{1} of 13 & UR-IW-2 & 1.0000 & 1.0000 & 1.0000 & \textbf{ 1.0000 } \\
\hline
\end{tabular}
\end{table}

\begin{table}[h!]
\caption{Task 12 Synergy, Q\&A exact Factoid, Batches 3-4}
\footnotesize
\label{tab:synergyFactoidBatches}
\begin{tabular}{|c|c|c|c|c|c|}
\hline
Batch & Position & System & Strict Acc. & Lenient Acc. & \textbf{MRR} \\
\hline
Test round 3 & 1 of 9 & Top Competitor & 0.4444 & 0.6667 & \textbf{ 0.5556 } \\
 & \textbf{4} of 9 & UR-IW-2 & 0.4444 & 0.5556 & \textbf{ 0.5000 } \\
 & \textbf{6} of 9 & UR-IW-3 & 0.4444 & 0.4444 & \textbf{ 0.4444 } \\
\hline
Test round 4 & \textbf{1} of 13 & \textbf{UR-IW-2} & 0.2727 & 0.6364 & \textbf{ 0.4318 } \\
 & \textbf{9} of 13 & UR-IW-3 & 0.2727 & 0.2727 & \textbf{ 0.2727 } \\
\hline
\end{tabular}
\end{table}

\begin{table}[h!]
\caption{Task 12 Synergy, Q\&A exact List, Batches 3-4}
\footnotesize
\label{tab:synergyListBatches}
\begin{tabular}{|c|c|c|c|c|c|}
\hline
Batch & Position & System & Mean Prec. & Recall & \textbf{F-Measure} \\
\hline
Test round 3 & 1 of 9 & Top Competitor & 0.4210 & 0.3280 & \textbf{ 0.3393 } \\
 & \textbf{7} of 9 & UR-IW-2 & 0.2500 & 0.3285 & \textbf{ 0.2669 } \\
 & \textbf{8} of 9 & UR-IW-3 & 0.2532 & 0.2710 & \textbf{ 0.2459 } \\
\hline
Test round 4 & 1 of 13 & Top Competitor & 0.3452 & 0.2794 & \textbf{ 0.2707 } \\
 & \textbf{4} of 13 & UR-IW-2 & 0.2054 & 0.3558 & \textbf{ 0.2395 } \\
 & \textbf{12} of 13 & UR-IW-3 & 0.1347 & 0.2203 & \textbf{ 0.1455 } \\
\hline
\end{tabular}
\end{table}

Completing one run with gpt-4-0125-preview cost around \$12 in API fees, while the same run with gpt-3.5-turbo-0125 was around 10 times cheaper at \$1.2. Gpt-4-0125-preview was also quite slow, taking around 180 minutes to complete one run, while gpt-3.5-turbo-0125 took only a few minutes. Cost significantly decreased compared to our last year's participation, while speed increased. This enabled us to actually use GPT-4 on the snippet extraction task, while last year we were not able to complete runs with snippet extraction for GPT-4 due to time and cost constraints. We also encountered less to none API errors during our runs, except for the empty responses in batch 4 for GPT-4, while last year we often had to rerun questions because the API timed-out or returned other errors.

\subsection{Task 12 B Phase A}
We participated with 5 systems in all 4 batches of Task 12 B Phase A. The systems either used 1- or 10-shot learning with the plain or a fine-tuned model in batches 1+2 or additional context retrieved from Wikipedia in batches 3-4. The system names and configurations are listed below.

\textbf{Batches 1-2:}
\begin{itemize}
    \item UR-IW-1: Claude 3 Opus + 1-shot
    \item UR-IW-2: Mixtral 8x7B Instruct v0.1, QloRa Fine-Tuned + 10-Shot
    \item UR-IW-3: gpt-3.5-turbo-0125 fine-tuned + 1-shot
    \item UR-IW-4: Mixtral 8x7B Instruct v0.1 + 10-shot
    \item UR-IW-5: gpt-3.5-turbo-0125 + 10-shot
\end{itemize}

\textbf{Batches 3-4:}
\begin{itemize}
    \item UR-IW-1: Claude 3 Opus 1-shot + wiki
    \item UR-IW-2: Claude 3 Opus 1-shot
    \item UR-IW-3: Mixtral 8x7B Instruct v0.1 10-Shot + wiki
    \item UR-IW-4: Mixtral 8x7B Instruct v0.1 10-Shot
    \item UR-IW-5: Mixtral 8x22B Instruct v0.1 10-Shot + wiki
\end{itemize}

The following Tables \ref{tab:12BPhaseADocuments} and \ref{tab:12BPhaseASnippets} show the results of our systems participating in the 4 batches. MAP was the official metric to compare the systems.

\begin{table}[h!]
\caption{Task 12B Phase A, Document Retrieval}
\footnotesize
\label{tab:12BPhaseADocuments}
\begin{tabular}{|c|c|c|c|c|c|c|c|}
\hline
Batch & Position & System & Precision & Recall & F-Measure & \textbf{MAP} & GMAP \\
\hline
Test Batch 1 & 1 of 40 & Top Competitor & 0.1039 & 0.3124 & 0.1485 & \textbf{ 0.2067 } & 0.0016 \\
 & \textbf{25} of 40 & UR-IW-5 & 0.0525 & 0.1093 & 0.0602 & \textbf{ 0.0811 } & 0.0001 \\
 & \textbf{26} of 40 & UR-IW-1 & 0.0784 & 0.1525 & 0.0938 & \textbf{ 0.0751 } & 0.0002 \\
 & \textbf{29} of 40 & UR-IW-3 & 0.0539 & 0.1023 & 0.0648 & \textbf{ 0.0631 } & 0.0001 \\
 & \textbf{30} of 40 & UR-IW-2 & 0.0544 & 0.0975 & 0.0551 & \textbf{ 0.0600 } & 0.0001 \\
 & \textbf{32} of 40 & UR-IW-4 & 0.0566 & 0.0861 & 0.0477 & \textbf{ 0.0511 } & 0.0001 \\
\hline
Test Batch 2 & 1 of 53 & Top Competitor & 0.0953 & 0.3673 & 0.1428 & \textbf{ 0.2293 } & 0.0026 \\
 & \textbf{32} of 53 & UR-IW-1 & 0.0889 & 0.1607 & 0.0971 & \textbf{ 0.0875 } & 0.0002 \\
 & \textbf{39} of 53 & UR-IW-2 & 0.0502 & 0.1227 & 0.0583 & \textbf{ 0.0657 } & 0.0001 \\
 & \textbf{40} of 53 & UR-IW-3 & 0.0542 & 0.1390 & 0.0716 & \textbf{ 0.0643 } & 0.0001 \\
 & \textbf{43} of 53 & UR-IW-5 & 0.0633 & 0.1048 & 0.0660 & \textbf{ 0.0564 } & 0.0001 \\
 & \textbf{45} of 53 & UR-IW-4 & 0.0694 & 0.0589 & 0.0517 & \textbf{ 0.0409 } & 0.0000 \\
\hline
Test Batch 3 & 1 of 58 & Top Competitor & 0.0859 & 0.3835 & 0.1309 & \textbf{ 0.2549 } & 0.0024 \\
 & \textbf{27} of 58 & UR-IW-1 & 0.0524 & 0.1761 & 0.0720 & \textbf{ 0.1281 } & 0.0002 \\
 & \textbf{31} of 58 & UR-IW-2 & 0.0541 & 0.1569 & 0.0734 & \textbf{ 0.1217 } & 0.0001 \\
 & \textbf{40} of 58 & UR-IW-3 & 0.0687 & 0.1730 & 0.0766 & \textbf{ 0.0971 } & 0.0002 \\
 & \textbf{42} of 58 & UR-IW-5 & 0.0664 & 0.1866 & 0.0854 & \textbf{ 0.0957 } & 0.0003 \\
 & \textbf{52} of 58 & UR-IW-4 & 0.0446 & 0.0859 & 0.0492 & \textbf{ 0.0480 } & 0.0001 \\
\hline
Test Batch 4 & 1 of 49 & Top Competitor & 0.1000 & 0.5569 & 0.1609 & \textbf{ 0.3930 } & 0.0148 \\
 & \textbf{17} of 49 & UR-IW-2 & 0.1199 & 0.3769 & 0.1586 & \textbf{ 0.2910 } & 0.0018 \\
 & \textbf{22} of 49 & UR-IW-1 & 0.0952 & 0.2810 & 0.1253 & \textbf{ 0.1892 } & 0.0006 \\
 & \textbf{23} of 49 & UR-IW-4 & 0.0934 & 0.2686 & 0.1224 & \textbf{ 0.1819 } & 0.0005 \\
 & \textbf{25} of 49 & UR-IW-5 & 0.0870 & 0.2231 & 0.1099 & \textbf{ 0.1617 } & 0.0003 \\
 & \textbf{35} of 49 & UR-IW-3 & 0.0681 & 0.1861 & 0.0844 & \textbf{ 0.1281 } & 0.0001 \\
\hline
\end{tabular}
\end{table}

In batches 1 \& 2 where we compared fine-tuned versions of GTP-3.5 and Mixtral 8x7b with 10-shot learning and Claude 3 Opus, no clear trend was observable over batches in the document retrieval stage of Phase A as can be seen in Table \ref{tab:12BPhaseADocuments}. Only Mixtral 8x7B with 10-shot learning was consistently performing worse than all our other models, While Claude 3 Opus with 1-shot learning was our best model in batch 2 and second best in batch 1. 

We used 1-shot learning instead of 10-shot learning for Claude 3 Opus due to time constraints because the model was slow, and for the fine-tuned gpt-3.5-turbo due to cost constraints. Sending 10 examples of abstracts for snippet extractions per 50 highest-ranked search results would have amounted to quite some input tokens per run for these models.

For batch 3 \& 4 where we explored if giving the systems additional Wikipedia context while creating queries for Elasticsearch could improve their performance, we also could not observe a consistent effect over batches. While in batch 3 the systems with additional Wikipedia context performed better, this effect was reversed in batch 4.

\begin{table}[h!]
\caption{Task 12B Phase A, Snippet Extraction}
\footnotesize
\label{tab:12BPhaseASnippets}
\begin{tabular}{|c|c|c|c|c|c|c|c|}
\hline
Batch & Position & System & Precision & Recall & F-Measure & \textbf{MAP} & GMAP \\
\hline
Test Batch 1 & 1 of 40 & Top Competitor & 0.0446 & 0.1490 & 0.0638 & \textbf{ 0.1149 } & 0.0001 \\
 & \textbf{7} of 40 & UR-IW-3 & 0.0454 & 0.0539 & 0.0458 & \textbf{ 0.0452 } & 0.0001 \\
 & \textbf{10} of 40 & UR-IW-5 & 0.0450 & 0.0546 & 0.0441 & \textbf{ 0.0412 } & 0.0000 \\
 & \textbf{11} of 40 & UR-IW-1 & 0.0480 & 0.0720 & 0.0508 & \textbf{ 0.0357 } & 0.0001 \\
 & \textbf{12} of 40 & UR-IW-4 & 0.0444 & 0.0527 & 0.0336 & \textbf{ 0.0244 } & 0.0001 \\
 & \textbf{13} of 40 & UR-IW-2 & 0.0341 & 0.0483 & 0.0276 & \textbf{ 0.0237 } & 0.0000 \\
\hline
Test Batch 2 & 1 of 53 & Top Competitor & 0.0520 & 0.1810 & 0.0746 & \textbf{ 0.1539 } & 0.0003 \\
 & \textbf{6} of 53 & UR-IW-1 & 0.0568 & 0.0850 & 0.0532 & \textbf{ 0.0569 } & 0.0001 \\
 & \textbf{11} of 53 & UR-IW-3 & 0.0400 & 0.0722 & 0.0474 & \textbf{ 0.0345 } & 0.0001 \\
 & \textbf{13} of 53 & UR-IW-5 & 0.0357 & 0.0474 & 0.0333 & \textbf{ 0.0301 } & 0.0000 \\
 & \textbf{17} of 53 & UR-IW-4 & 0.0590 & 0.0449 & 0.0334 & \textbf{ 0.0230 } & 0.0000 \\
 & \textbf{18} of 53 & UR-IW-2 & 0.0329 & 0.0713 & 0.0278 & \textbf{ 0.0191 } & 0.0000 \\
\hline
Test Batch 3 & 1 of 58 & Top Competitor & 0.0666 & 0.2568 & 0.0940 & \textbf{ 0.2224 } & 0.0009 \\
 & \textbf{6} of 58 & UR-IW-1 & 0.0379 & 0.1251 & 0.0508 & \textbf{ 0.0818 } & 0.0002 \\
 & \textbf{7} of 58 & UR-IW-5 & 0.0399 & 0.1188 & 0.0506 & \textbf{ 0.0736 } & 0.0002 \\
 & \textbf{8} of 58 & UR-IW-2 & 0.0359 & 0.0881 & 0.0456 & \textbf{ 0.0677 } & 0.0001 \\
 & \textbf{20} of 58 & UR-IW-3 & 0.0320 & 0.0819 & 0.0338 & \textbf{ 0.0402 } & 0.0001 \\
 & \textbf{21} of 58 & UR-IW-4 & 0.0316 & 0.0388 & 0.0279 & \textbf{ 0.0381 } & 0.0000 \\
\hline
Test Batch 4 & 1 of 49 & Top Competitor & 0.0782 & 0.4162 & 0.1191 & \textbf{ 0.3437 } & 0.0043 \\
 & \textbf{6} of 49 & UR-IW-2 & 0.0777 & 0.1846 & 0.0888 & \textbf{ 0.1402 } & 0.0008 \\
 & \textbf{8} of 49 & UR-IW-1 & 0.0502 & 0.1398 & 0.0645 & \textbf{ 0.0959 } & 0.0002 \\
 & \textbf{10} of 49 & UR-IW-4 & 0.0559 & 0.0848 & 0.0566 & \textbf{ 0.0661 } & 0.0001 \\
 & \textbf{11} of 49 & UR-IW-5 & 0.0586 & 0.1208 & 0.0654 & \textbf{ 0.0617 } & 0.0001 \\
 & \textbf{14} of 49 & UR-IW-3 & 0.0400 & 0.0486 & 0.0329 & \textbf{ 0.0428 } & 0.0000 \\
\hline
\end{tabular}
\end{table}

In the snippet extraction stage of Phase A, our QloRa fine-tuned version of Mixtral 8x7B was our worst-performing system in both batch 1 \& 2, followed by the 10-shot Mixtral 8x7B version as can be seen in Table \ref{tab:12BPhaseASnippets}. The fine-tuned version of GPT-3.5-turbo was consistently ahead of its 10-shot counterpart in both batches, while Claude 3 Opus was worse than the GPT-3.4-turbo systems in batch 1 and better in batch 2.

Additional Wikipedia context did not lead to consistent results across batches. While the systems with Wikipedia context (UR-IW-1, UR-IW-3) performed better than their counterparts (UR-IW-2, UR-IW-4) in batch 3, this effect was again reversed in batch 4.

One run with Claude 3 Opus with 1-shot learning and additional Wikipedia context took around 140 minutes to complete, we did not have to pay for the tokens used because we had an early beta evaluation account. For Mixtral 8x7B with 10-shot learning and additional Wikipedia context, the runs took around 14 minutes to complete via the fireworks.ai API and cost around \$ 11. Fireworks.ai charged \$0.50 /1M tokens for both input and output tokens as of writing, while Anthropic would have charged \$ 15 /1M tokens for input and \$ 75 /1M tokens for output. So the cost for doing 10-shot learning with Claude 3 Opus would have been at least 30 times as high while being 10 times slower. 

\subsection{Task 12B Phase A+}
We participated with 5 systems in nearly all 4 batches of Task 12 B Phase A+\footnote{We failed to submit one system run for system number 5 in batch 3.}. The systems either used 10-shot learning with the plain or a fine-tuned model in batches 1+2 or additional context retrieved from Wikipedia in batches 3-4. Per batch, we used the same input snippet file for all systems to base their answers on to ensure that their performance is comparable.

\textbf{Batches 1-2:}
\begin{itemize}
    \item UR-IW-1: Claude 3 Opus + 10-shot
    \item UR-IW-2: Mixtral 8x7B Instruct v0.1, QLoRa Fine-Tuned + 10-Shot
    \item UR-IW-3: gpt-3.5-turbo-0125 fine-tuned + 10-shot
    \item UR-IW-4: Mixtral 8x7B Instruct v0.1 + 10-shot
    \item UR-IW-5: gpt-3.5-turbo-0125 + 10-shot
\end{itemize}

\textbf{Batches 3-4:}
\begin{itemize}
    \item UR-IW-1: Claude 3 Opus 10-shot + wiki
    \item UR-IW-2: Mixtral 8x22B Instruct v0.1 10-Shot
    \item UR-IW-3: Mixtral 8x7B Instruct v0.1 10-Shot + wiki
    \item UR-IW-4: Mixtral 8x7B Instruct v0.1 10-Shot
    \item UR-IW-5: Mixtral 8x22B Instruct v0.1 10-Shot + wiki
\end{itemize}

\begin{table}[h!]
\caption{Task 12B Phase A+, exact questions Yes/No}
\footnotesize
\label{tab:12BPhaseAPlusYesNo}
\begin{tabular}{|c|c|c|c|c|c|c|}
\hline
Batch & Position & System & Accuracy & F1 Yes & F1 No & \textbf{Macro F1} \\
\hline
Test Batch 1 & \textbf{1} of 22 & \textbf{UR-IW-3} & 0.9200 & 0.9333 & 0.9000 & \textbf{ 0.9167 } \\
 & \textbf{4} of 22 & UR-IW-4 & 0.8400 & 0.8462 & 0.8333 & \textbf{ 0.8397 } \\
 & \textbf{5} of 22 & UR-IW-2 & 0.8400 & 0.8462 & 0.8333 & \textbf{ 0.8397 } \\
 & \textbf{6} of 22 & UR-IW-5 & 0.8000 & 0.8148 & 0.7826 & \textbf{ 0.7987 } \\
 & \textbf{8} of 22 & UR-IW-1 & 0.8000 & 0.8276 & 0.7619 & \textbf{ 0.7947 } \\
\hline
Test Batch 2 & 1 of 26 & Top Competitor & 0.9615 & 0.9677 & 0.9524 & \textbf{ 0.9601 } \\
 & \textbf{2} of 26 & UR-IW-5 & 0.8846 & 0.8966 & 0.8696 & \textbf{ 0.8831 } \\
 & \textbf{3} of 26 & UR-IW-3 & 0.8846 & 0.8966 & 0.8696 & \textbf{ 0.8831 } \\
 & \textbf{6} of 26 & UR-IW-4 & 0.8462 & 0.8571 & 0.8333 & \textbf{ 0.8452 } \\
 & \textbf{7} of 26 & UR-IW-2 & 0.8462 & 0.8571 & 0.8333 & \textbf{ 0.8452 } \\
 & \textbf{12} of 26 & UR-IW-1 & 0.7692 & 0.8000 & 0.7273 & \textbf{ 0.7636 } \\
\hline
Test Batch 3 & \textbf{1} of 28 & \textbf{UR-IW-5} & 0.9167 & 0.9286 & 0.9000 & \textbf{ 0.9143 } \\
 & \textbf{8} of 28 & UR-IW-2 & 0.8333 & 0.8462 & 0.8182 & \textbf{ 0.8322 } \\
 & \textbf{10} of 28 & UR-IW-1 & 0.8333 & 0.8667 & 0.7778 & \textbf{ 0.8222 } \\
 & \textbf{11} of 28 & UR-IW-3 & 0.7917 & 0.8000 & 0.7826 & \textbf{ 0.7913 } \\
 & \textbf{13} of 28 & UR-IW-4 & 0.7917 & 0.8148 & 0.7619 & \textbf{ 0.7884 } \\
\hline
Test Batch 4 & 1 of 29 & Top Competitor & 0.8889 & 0.9189 & 0.8235 & \textbf{ 0.8712 } \\
 & \textbf{3} of 29 & UR-IW-1 & 0.8519 & 0.8947 & 0.7500 & \textbf{ 0.8224 } \\
 & \textbf{4} of 29 & UR-IW-2 & 0.8519 & 0.8947 & 0.7500 & \textbf{ 0.8224 } \\
 & \textbf{9} of 29 & UR-IW-4 & 0.7778 & 0.8333 & 0.6667 & \textbf{ 0.7500 } \\
 & \textbf{11} of 29 & UR-IW-5 & 0.7407 & 0.8000 & 0.6316 & \textbf{ 0.7158 } \\
 & \textbf{14} of 29 & UR-IW-3 & 0.7037 & 0.7647 & 0.6000 & \textbf{ 0.6824 } \\
\hline
\end{tabular}
\end{table}

For the yes/no exact answer format, Claude 3 Opus with 10-shot learning was our worst-performing system, while the fine-tuned version of GPT-3.5-turbo was our top-performing system in batch 1 and only beaten by its 10-shot counter-part in batch 2, as can be seen in Table \ref{tab:12BPhaseAPlusYesNo}. \textbf{It was interesting to see that the open-source models could perform better than the presumably most advanced commercial model, Claude 3 Opus, in this task}.

For batches 3 \& 4, we could show that additional Wikipedia context led to inconsistent results. While this context improved performance in batch 3 for the Wikipedia enhanced systems (UR-IW-5, UR-IW-3) over their normal 10-shot counterparts (UR-IW-2, UR-IW-4) it again led to worse performance in batch 4. This result is in line with the results from Phase A where these systems performed similarly for document retrieval and snippet extraction. We speculate that the models are sensitive to the Wikipedia context, and the usefulness of the context is highly influenced by both the entities present in the questions, and its relationship to the relevant snippets. 

\begin{table}[h!]
\caption{Task 12 B, Phase A+, exact questions factoid}
\footnotesize
\label{tab:12BPhaseAPlusFactoid}
\begin{tabular}{|c|c|c|c|c|c|}
\hline
Batch & Position & System & Strict Acc. & Lenient Acc. & \textbf{MRR} \\
\hline
Test Batch 1 & 1 of 22 & Top Competitor & 0.2381 & 0.5238 & \textbf{ 0.3611 } \\
 & \textbf{4} of 22 & UR-IW-1 & 0.1905 & 0.2381 & \textbf{ 0.2143 } \\
 & \textbf{10} of 22 & UR-IW-5 & 0.0952 & 0.0952 & \textbf{ 0.0952 } \\
 & \textbf{12} of 22 & UR-IW-2 & 0.0952 & 0.0952 & \textbf{ 0.0952 } \\
 & \textbf{13} of 22 & UR-IW-3 & 0.0952 & 0.0952 & \textbf{ 0.0952 } \\
 & \textbf{14} of 22 & UR-IW-4 & 0.0476 & 0.0952 & \textbf{ 0.0714 } \\
\hline
Test Batch 2 & 1 of 26 & Top Competitor & 0.3684 & 0.4211 & \textbf{ 0.3947 } \\
 & \textbf{3} of 26 & UR-IW-5 & 0.3158 & 0.3158 & \textbf{ 0.3158 } \\
 & \textbf{4} of 26 & UR-IW-3 & 0.3158 & 0.3158 & \textbf{ 0.3158 } \\
 & \textbf{7} of 26 & UR-IW-2 & 0.2632 & 0.3158 & \textbf{ 0.2895 } \\
 & \textbf{9} of 26 & UR-IW-1 & 0.2632 & 0.2632 & \textbf{ 0.2632 } \\
 & \textbf{14} of 26 & UR-IW-4 & 0.1579 & 0.2105 & \textbf{ 0.1842 } \\
\hline
Test Batch 3 & 1 of 28 & Top Competitor & 0.2692 & 0.4231 & \textbf{ 0.3301 } \\
 & \textbf{8} of 28 & UR-IW-1 & 0.1923 & 0.3077 & \textbf{ 0.2340 } \\
 & \textbf{10} of 28 & UR-IW-4 & 0.1923 & 0.2308 & \textbf{ 0.2019 } \\
 & \textbf{11} of 28 & UR-IW-2 & 0.1538 & 0.1923 & \textbf{ 0.1731 } \\
 & \textbf{16} of 28 & UR-IW-3 & 0.1538 & 0.1538 & \textbf{ 0.1538 } \\
 & \textbf{17} of 28 & UR-IW-5 & 0.1538 & 0.1538 & \textbf{ 0.1538 } \\
\hline
Test Batch 4 & 1 of 29 & Top Competitor & 0.3684 & 0.4211 & \textbf{ 0.3947 } \\
 & \textbf{2} of 29 & UR-IW-1 & 0.3158 & 0.4737 & \textbf{ 0.3816 } \\
 & \textbf{3} of 29 & UR-IW-5 & 0.3684 & 0.3684 & \textbf{ 0.3684 } \\
 & \textbf{4} of 29 & UR-IW-2 & 0.3158 & 0.3684 & \textbf{ 0.3421 } \\
 & \textbf{8} of 29 & UR-IW-3 & 0.2105 & 0.3158 & \textbf{ 0.2412 } \\
 & \textbf{10} of 29 & UR-IW-4 & 0.1579 & 0.2632 & \textbf{ 0.2018 } \\
\hline
\end{tabular}
\end{table}

In the exact answer factoid format, our worst-performing system in batches 1 \& 2 was consistently Mixtral 8x7B with 10-shot learning while its fine-tuned counterpart performed better as can be seen in Table \ref{tab:12BPhaseAPlusFactoid}. This order was reversed for GPT-3.5-turbo, where the fine-tuned version performed worse than its counterpart.

The additional Wikipedia context again led to inconsistent results across batches, but this time the systems with Wikipedia context performed better in batch 4 compared to batch 3, which is contrary to the observed behavior in the document retrieval and snippet extraction in Phase A as well as the yes/no answer format in Phase A+.

\begin{table}[h!]
\caption{Task 12 B, Phase A+, exact questions list}
\footnotesize
\label{tab:12BPhaseAPlusList}
\begin{tabular}{|c|c|c|c|c|c|}
\hline
Batch & Position & System & Mean Prec. & Recall & \textbf{F-Measure} \\
\hline
Test Batch 1 & \textbf{1} of 22 & \textbf{UR-IW-2} & 0.5250 & 0.4914 & \textbf{ 0.4808 } \\
 & \textbf{3} of 22 & UR-IW-3 & 0.4016 & 0.4778 & \textbf{ 0.4089 } \\
 & \textbf{4} of 22 & UR-IW-5 & 0.4119 & 0.4182 & \textbf{ 0.3976 } \\
 & \textbf{5} of 22 & UR-IW-4 & 0.3948 & 0.4063 & \textbf{ 0.3798 } \\
 & \textbf{7} of 22 & UR-IW-1 & 0.3224 & 0.4273 & \textbf{ 0.3418 } \\
\hline
Test Batch 2 & 1 of 26 & Top Competitor & 0.4470 & 0.4451 & \textbf{ 0.4088 } \\
 & \textbf{7} of 26 & UR-IW-3 & 0.2625 & 0.2400 & \textbf{ 0.2411 } \\
 & \textbf{8} of 26 & UR-IW-2 & 0.2045 & 0.2569 & \textbf{ 0.2182 } \\
 & \textbf{9} of 26 & UR-IW-4 & 0.2628 & 0.2299 & \textbf{ 0.2179 } \\
 & \textbf{12} of 26 & UR-IW-1 & 0.1953 & 0.1906 & \textbf{ 0.1766 } \\
 & \textbf{14} of 26 & UR-IW-5 & 0.1589 & 0.1725 & \textbf{ 0.1497 } \\
\hline
Test Batch 3 & 1 of 28 & Top Competitor & 0.3750 & 0.4069 & \textbf{ 0.3708 } \\
 & \textbf{4} of 28 & UR-IW-1 & 0.2657 & 0.4232 & \textbf{ 0.3000 } \\
 & \textbf{10} of 28 & UR-IW-5 & 0.2208 & 0.2881 & \textbf{ 0.2392 } \\
 & \textbf{12} of 28 & UR-IW-2 & 0.2125 & 0.2892 & \textbf{ 0.2303 } \\
 & \textbf{13} of 28 & UR-IW-4 & 0.2014 & 0.2655 & \textbf{ 0.2186 } \\
 & \textbf{19} of 28 & UR-IW-3 & 0.1373 & 0.2326 & \textbf{ 0.1627 } \\
\hline
Test Batch 4 & 1 of 29 & Top Competitor & 0.3139 & 0.3433 & \textbf{ 0.3219 } \\
 & \textbf{8} of 29 & UR-IW-1 & 0.1529 & 0.2641 & \textbf{ 0.1774 } \\
 & \textbf{14} of 29 & UR-IW-4 & 0.1364 & 0.1845 & \textbf{ 0.1418 } \\
 & \textbf{15} of 29 & UR-IW-2 & 0.1161 & 0.2069 & \textbf{ 0.1366 } \\
 & \textbf{17} of 29 & UR-IW-5 & 0.1125 & 0.1610 & \textbf{ 0.1269 } \\
 & \textbf{18} of 29 & UR-IW-3 & 0.1191 & 0.1239 & \textbf{ 0.1155 } \\
\hline
\end{tabular}
\end{table}

For the list exact answer format, Mixtral 8x7B was again our worst-performing system in batch 1 \& 2, while its fine-tuned counterpart was competing with the fine-tuned version of gpt-3.5-turbo for the top positions as can be seen in Table \ref{tab:12BPhaseAPlusList}. 

In batches 3 \& 4 Claude 3 Opus with 10-shot learning and additional Wikipedia context was the best-performing system in both batches while Mixtral 8x7B with 10-shot learning and additional Wikipedia context was the worst-performing one. \textbf{Overall no consistent effect of the Wikipedia context was observable across models and batches}.

\subsection{Task 12B Phase B}
We participated with 5 systems in all 4 batches of Task 12B Phase B. The systems used 10-shot learning with the plain or a fine-tuned model in batches 1 \& 2 or additional context retrieved from Wikipedia in batches 3 \& 4.

\textbf{Batches 1-2:}
\begin{itemize}
    \item UR-IW-1: Claude 3 Opus + 10-shot
    \item UR-IW-2: Mixtral 8x7B Instruct v0.1, QLoRa Fine-Tuned + 10-Shot
    \item UR-IW-3: gpt-3.5-turbo-0125 fine-tuned + 10-shot
    \item UR-IW-4: Mixtral 8x7B Instruct v0.1 + 10-shot
    \item UR-IW-5: gpt-3.5-turbo-0125 + 10-shot
\end{itemize}

\textbf{Batch 3}
\begin{itemize}
    \item UR-IW-1: Claude 3 Opus 10-shot + wiki
    \item UR-IW-2: Mixtral 8x22B Instruct v0.1 10-Shot
    \item UR-IW-3: Mixtral 8x7B Instruct v0.1 10-Shot + wiki
    \item UR-IW-4: Mixtral 8x7B Instruct v0.1 10-Shot
    \item UR-IW-5: Mixtral 8x22B Instruct v0.1 10-Shot + wiki
\end{itemize}

\textbf{Batch 4}
\begin{itemize}
    \item UR-IW-1: Claude 3 Opus 10-shot + wiki
    \item UR-IW-2: Claude 3 Opus 10-shot
    \item UR-IW-3: Mixtral 8x7B Instruct v0.1 10-Shot + wiki
    \item UR-IW-4: Mixtral 8x7B Instruct v0.1 10-Shot
    \item UR-IW-5: Mixtral 8x22B Instruct v0.1 10-Shot + wiki
\end{itemize}

\begin{table}[h!]
\caption{Task 12 B, Phase B, exact Yes/No}
\footnotesize
\label{tab:12BPhaseBYesNo}
\begin{tabular}{|c|c|c|c|c|c|c|}
\hline
Batch & Position & System & Accuracy & F1 Yes & F1 No & \textbf{Macro F1} \\
\hline
Test Batch 1 & \textbf{1} of 39 & \textbf{UR-IW-1} & 0.9600 & 0.9655 & 0.9524 & \textbf{ 0.9589 } \\
 & \textbf{2} of 39 & \textbf{UR-IW-5} & 0.9600 & 0.9655 & 0.9524 & \textbf{ 0.9589 } \\
 & \textbf{5} of 39 & UR-IW-2 & 0.9200 & 0.9231 & 0.9167 & \textbf{ 0.9199 } \\
 & \textbf{6} of 39 & UR-IW-4 & 0.9200 & 0.9286 & 0.9091 & \textbf{ 0.9188 } \\
 & \textbf{7} of 39 & UR-IW-3 & 0.9200 & 0.9286 & 0.9091 & \textbf{ 0.9188 } \\
\hline
Test Batch 2 & \textbf{1} of 43 & \textbf{UR-IW-3} & 0.9615 & 0.9677 & 0.9524 & \textbf{ 0.9601 } \\
 & \textbf{4} of 43 & UR-IW-1 & 0.9615 & 0.9697 & 0.9474 & \textbf{ 0.9585 } \\
 & \textbf{8} of 43 & UR-IW-2 & 0.9231 & 0.9375 & 0.9000 & \textbf{ 0.9188 } \\
 & \textbf{9} of 43 & UR-IW-5 & 0.9231 & 0.9375 & 0.9000 & \textbf{ 0.9188 } \\
 & \textbf{26} of 43 & UR-IW-4 & 0.8462 & 0.8667 & 0.8182 & \textbf{ 0.8424 } \\
\hline
Test Batch 3 & 1 of 48 & Top Competitor & 1.0000 & 1.0000 & 1.0000 & \textbf{ 1.0000 } \\
 & \textbf{17} of 48 & UR-IW-1 & 0.9167 & 0.9286 & 0.9000 & \textbf{ 0.9143 } \\
 & \textbf{23} of 48 & UR-IW-2 & 0.8750 & 0.8800 & 0.8696 & \textbf{ 0.8748 } \\
 & \textbf{26} of 48 & UR-IW-3 & 0.8750 & 0.8889 & 0.8571 & \textbf{ 0.8730 } \\
 & \textbf{27} of 48 & UR-IW-4 & 0.8750 & 0.8889 & 0.8571 & \textbf{ 0.8730 } \\
\hline
Test Batch 4 & 1 of 49 & Top Competitor & 0.9630 & 0.9730 & 0.9412 & \textbf{ 0.9571 } \\
 & \textbf{8} of 49 & UR-IW-1 & 0.9259 & 0.9444 & 0.8889 & \textbf{ 0.9167 } \\
 & \textbf{19} of 49 & UR-IW-2 & 0.8889 & 0.9231 & 0.8000 & \textbf{ 0.8615 } \\
 & \textbf{20} of 49 & UR-IW-4 & 0.8519 & 0.8889 & 0.7778 & \textbf{ 0.8333 } \\
 & \textbf{25} of 49 & UR-IW-5 & 0.8148 & 0.8571 & 0.7368 & \textbf{ 0.7970 } \\
 & \textbf{31} of 49 & UR-IW-3 & 0.5926 & 0.5926 & 0.5926 & \textbf{ 0.5926 } \\
\hline
\end{tabular}
\end{table}

In the exact yes/no answer settings of Phase B, \textbf{Claude 3 Opus with 10-shot learning and gpt-3.5 turbo with 10-shot learning were sharing first place in batch 1} while the fine-tuned version of gpt-3.5-turbo was the best-performing system in batch 2 as can be seen in Table \ref{tab:12BPhaseBYesNo}. The fine-tuned version of Mixtral 8x7B was also competitive, taking the 5th position in batch 1 and 8th position in batch 2. 

In batch 3 \& 4 with additional wikipedia context the systems performed better than their counterparts in batch 3 while this result was mixed in batch 4, leading again to inconsistent results.

\begin{table}[h!]
\caption{Task 12B, Phase B, exact factoid}
\footnotesize
\label{tab:12BPhaseBFactoid}
\begin{tabular}{|c|c|c|c|c|c|}
\hline
Batch & Position & System & Strict Acc. & Lenient Acc. & \textbf{MRR} \\
\hline
Test Batch 1 & 1 of 39 & Top Competitor & 0.4286 & 0.4286 & \textbf{ 0.4286 } \\
 & \textbf{11} of 39 & UR-IW-1 & 0.1905 & 0.3333 & \textbf{ 0.2540 } \\
 & \textbf{12} of 39 & UR-IW-5 & 0.2381 & 0.2857 & \textbf{ 0.2540 } \\
 & \textbf{14} of 39 & UR-IW-2 & 0.2381 & 0.2381 & \textbf{ 0.2381 } \\
 & \textbf{15} of 39 & UR-IW-3 & 0.2381 & 0.2381 & \textbf{ 0.2381 } \\
 & \textbf{23} of 39 & UR-IW-4 & 0.1905 & 0.1905 & \textbf{ 0.1905 } \\
\hline
Test Batch 2 & \textbf{1} of 43 & \textbf{UR-IW-1} & 0.6316 & 0.7368 & \textbf{ 0.6842 } \\
 & \textbf{2} of 43 & UR-IW-2 & 0.6842 & 0.6842 & \textbf{ 0.6842 } \\
 & \textbf{3} of 43 & UR-IW-4 & 0.6316 & 0.6316 & \textbf{ 0.6316 } \\
 & \textbf{8} of 43 & UR-IW-3 & 0.5263 & 0.5263 & \textbf{ 0.5263 } \\
 & \textbf{14} of 43 & UR-IW-5 & 0.4211 & 0.4211 & \textbf{ 0.4211 } \\
\hline
Test Batch 3 & 1 of 48 & Top Competitor & 0.5000 & 0.5000 & \textbf{ 0.5000 } \\
 & \textbf{7} of 48 & UR-IW-2 & 0.3846 & 0.3846 & \textbf{ 0.3846 } \\
 & \textbf{8} of 48 & UR-IW-3 & 0.3462 & 0.4231 & \textbf{ 0.3846 } \\
 & \textbf{13} of 48 & UR-IW-4 & 0.3462 & 0.3846 & \textbf{ 0.3654 } \\
 & \textbf{26} of 48 & UR-IW-1 & 0.2692 & 0.3077 & \textbf{ 0.2885 } \\
\hline
Test Batch 4 & \textbf{1} of 49 & \textbf{UR-IW-2} & 0.6316 & 0.6842 & \textbf{ 0.6579 } \\
 & \textbf{5} of 49 & UR-IW-5 & 0.5789 & 0.5789 & \textbf{ 0.5789 } \\
 & \textbf{12} of 49 & UR-IW-1 & 0.4737 & 0.6316 & \textbf{ 0.5439 } \\
 & \textbf{15} of 49 & UR-IW-4 & 0.4737 & 0.5789 & \textbf{ 0.5175 } \\
 & \textbf{27} of 49 & UR-IW-3 & 0.3684 & 0.3684 & \textbf{ 0.3684 } \\
\hline
\end{tabular}
\end{table}

For the exact factoid answer format in Phase B, \textbf{Claude 3 Opus and our fine-tuned Mixtral 8x7B model where sharing first place in batch 2} while in batch 3, Claude 3 Opus and gpt-3.5-turbo where the on the same level as can be seen in Table \ref{tab:12BPhaseBFactoid}.

For Mixtral 8x7B, additional Wikipedia context improved the outcome in batch 3 but led to worse results in batch 4, while a similar effect was observable for Claude 3 opus in batch 4\footnote{The run of Mixtral 8x22B with Wikipedia context was not successfully submitted to batch 3, we might have overlooked to upload them.}.

\begin{table}[h!]
\caption{Task 12 B, Phase B, exact List}
\footnotesize
\label{tab:12BPhaseBList}
\begin{tabular}{|c|c|c|c|c|c|}
\hline
Batch & Position & System & Mean Prec. & Recall & \textbf{F-Measure} \\
\hline
Test Batch 1 & 1 of 39 & Top Competitor & 0.6647 & 0.5804 & \textbf{ 0.5843 } \\
 & \textbf{3} of 39 & UR-IW-5 & 0.6054 & 0.5942 & \textbf{ 0.5790 } \\
 & \textbf{5} of 39 & UR-IW-3 & 0.6010 & 0.5799 & \textbf{ 0.5656 } \\
 & \textbf{13} of 39 & UR-IW-2 & 0.5202 & 0.4947 & \textbf{ 0.4992 } \\
 & \textbf{18} of 39 & UR-IW-1 & 0.4840 & 0.5069 & \textbf{ 0.4662 } \\
 & \textbf{22} of 39 & UR-IW-4 & 0.4563 & 0.3903 & \textbf{ 0.4015 } \\
\hline
Test Batch 2 & \textbf{1} of 43 & \textbf{UR-IW-4} & 0.5863 & 0.5645 & \textbf{ 0.5708 } \\
 & \textbf{2} of 43 & UR-IW-2 & 0.5835 & 0.5645 & \textbf{ 0.5698 } \\
 & \textbf{4} of 43 & UR-IW-3 & 0.5650 & 0.5347 & \textbf{ 0.5434 } \\
 & \textbf{8} of 43 & UR-IW-1 & 0.5061 & 0.5246 & \textbf{ 0.5047 } \\
 & \textbf{9} of 43 & UR-IW-5 & 0.5009 & 0.5347 & \textbf{ 0.5033 } \\
\hline
Test Batch 3 & 1 of 48 & Top Competitor & 0.6466 & 0.6560 & \textbf{ 0.6484 } \\
 & \textbf{3} of 48 & UR-IW-4 & 0.5656 & 0.5696 & \textbf{ 0.5611 } \\
 & \textbf{8} of 48 & UR-IW-2 & 0.5031 & 0.5367 & \textbf{ 0.5093 } \\
 & \textbf{15} of 48 & UR-IW-3 & 0.4451 & 0.4578 & \textbf{ 0.4473 } \\
 & \textbf{20} of 48 & UR-IW-1 & 0.3476 & 0.6010 & \textbf{ 0.4118 } \\
\hline
Test Batch 4 & 1 of 49 & Top Competitor & 0.7680 & 0.6266 & \textbf{ 0.6637 } \\
 & \textbf{7} of 49 & UR-IW-2 & 0.6209 & 0.6612 & \textbf{ 0.6299 } \\
 & \textbf{16} of 49 & UR-IW-5 & 0.5097 & 0.5044 & \textbf{ 0.4989 } \\
 & \textbf{18} of 49 & UR-IW-4 & 0.4919 & 0.4839 & \textbf{ 0.4699 } \\
 & \textbf{19} of 49 & UR-IW-3 & 0.4641 & 0.5067 & \textbf{ 0.4667 } \\
 & \textbf{22} of 49 & UR-IW-1 & 0.3634 & 0.5532 & \textbf{ 0.4260 } \\
\hline
\end{tabular}
\end{table}

For the exact list answer format in Phase B, Mixtral 8x7B with 10-shot learning took first place in batch 2 while being on place 22 of 39 in batch 1 where gpt-3-5-turbo with 10-shot learning was our best-performing one as can be seen in Table \ref{tab:12BPhaseBList}. 

For batches 3 \& 4 both Claude 3 Opus and Mixtral 8x7B performed worse with additional Wikipedia context across batches. In batch 3 our best-performing system was Mixtral 8x7B with 10-shot learning and in batch 4 it was Claude 3 Opus with 10-shot learning.

The costs for completing runs in Phase B were lower than in Phase A and the runs were faster because we did not have to do snippet extraction for 50 documents per question, times the number of few-shot examples, we therefore were able to also do 10-shot learning with Claude 3 Opus and the fine-tuned version of GPT-3.5-turbo. The processing with Mixtral 8x7B via the Fireworks.ai API only took around 30 seconds for plain 10-shot examples and around 2 minutes for 10-shot examples and additional Wikipedia context.

We also submitted ideal answers for Task B and A+, but do not report on the preliminary results here, as the official judging metric for this answer type is based on the manual judgements that are not available yet.

\section{Discussion and Future Work}\label{Discussion}
While testing both commercial and open-source models, we could observe that there was no clear dominating model across batches or sub-tasks. Even our presumably weakest model, Mixtral 8x7B Instruct v0.1 with 10-shot learning was able to secure some leading spots in some batches of the competition, beating all other competing systems, (see batch 3 in Table \ref{tab:12BPhaseAPlusYesNo} and batch 2 in Table \ref{tab:12BPhaseBList}). We speculate that both the RAG setting and 10-shot learning might level the playing field a bit between commercial and open-source models, and it indicates that there is clear potential for creating state-of-the-art systems even with cheaper, faster and presumably smaller open-source models, if they are used in the right way.

While the Mixtral model weights are publicly available, their training data is not published, which makes these models not ideal candidates for scientific research. A truly open-source LLM alternative is OLMo \cite{groeneveld2024olmo} published by the Allen Institute for AI. We choose Mixtral nevertheless because we wanted to study models that might be used by commercial practitioners in clinical or enterprise use cases, and we think the permissive license combined with the seemingly competitive performance on public benchmarks and its large context length makes it an ideal candidate for these use cases.

From our results with our experiments with additional Wikipedia context in BioASQ we could see that it had an impact on performance, but it was inconsistent across question batches. For some questions in some subtasks it led to improvements, while for others the performance declined. We speculate that this might be dependent on the relevant entities in the questions and the quality of the retrieved Wikipedia context. Further experiments are needed to analyze the impact of this additional context. 

We also speculate that Wikipedia might not be a good proxy knowledge base for doing domain-specific RAG for these models because they are probably already highly trained on Wikipedia data and therefore the additional knowledge from this source might not tell the models much that they not already know.

Another reason for the inconsistent results with Wikipedia context could be that we only prepended the context for the last prompt, and the preceding n-shot examples were not generated with taking this context into account. 

Regarding fine-tuning, we had the impression that the commercial offering from OpenAI was not worth the cost. Even though it led to top results in some batches (batch 3 in Table \ref{tab:12BPhaseBYesNo}) it also produced models with worse results than their significantly cheaper non fine-tuned counterpart in others. Maybe with the right training set and the right training run you might get a consistently superior model, but then you also have to add the engineering cost compared to simple 10-shot learning to the already more expensive usage and fine-tuning costs.

We had a similar impression regarding the QLoRa fine-tuning that we explored for Mixtral 8x7B. For example, the Mixtral model fine-tuned for list question answering was performing better than all other systems in batch 1 of Phase A+ (see Table \ref{tab:12BPhaseAPlusList}) but worse than its non fine-tuned counterpart in batch 2 of Phase B (see Table \ref{tab:12BPhaseBList}). Overall, adapter fine-tuning appears not to be straightforward and requires more time for dataset creation, training and testing than simple few-shot learning. A more promising research direction might be selecting optimal few-shot examples for a given task.

It is important to note that most of the results we presented here are preliminary and might change when the manual assessment of the system responses is completed by the BioASQ experts. But we expect that yes/no answers will stay the same and factoid and list answers might just change slightly.

The preliminary performance of our systems in the document retrieval stage (Phase A) was quite poor compared to the other systems. We speculate that our approach of relying on TF\_IDF-based retrieval and adding a richer semantic representation to the keyword query instead of using embeddings and vector search to add such information is not in line with the baseline system used to create the preliminary gold set. If that is true, it might be possible that our retrieval performance is actually better than we expect and the performance improves when the final results are out. But it could also just be that the approach is inferior. The good performance of our systems in Phase A+ where the questions had to be answered without gold snippets, might indicate that our retrieved snippets are not as useless as the preliminary results from Phase A suggest.

For future work, we would like to further explore optimal few-shot example selection \cite{lu2022fantastically}\cite{zhao2021calibrate}\cite{s2024prompt}, as few-shot learning seems to offer the best flexibility and requires less engineering effort compared to fine-tuning while being transferable between models. We also would like to revisit our knowledge base context augmentation approach, beyond the BioASQ challenge. We think that on a more technical test set where the relevant knowledge is highly unlikely to be present in the pre-training of these models, this approach could have a bigger impact. We also would need to use a different knowledge source as the English Wikipedia, which is part of most pre-training datasets \cite{gao2020pile800gbdatasetdiverse}.

\section{Ethical Considerations}\label{Ethics}
The current generation of LLMs still exhibit the phenomenon of so-called hallucinations \cite{ji2023survey} that is they sometimes make up factually incorrect statements and even harmful misinformation. LLMs might also reproduce myths or misinformation that they encountered during their open-domain training or that might have been added to their input context. A recent prominent case was Google's new AI overview feature suggesting users should add glue to their pizza\footnote{\url{https://web.archive.org/web/20240529100801/https://www.theverge.com/2024/5/23/24162896/google-ai-overview-hallucinations-glue-in-pizza}}. 

The hallucination and misinformation problems seem fundamental and difficult to solve as they have been known for quite some time now and even Google, one of the most experienced AI research companies, was unable to save itself from repeated embarrassment.

Even though RAG has been shown to reduce hallucinations in some settings \cite{shuster2021retrieval} occasional hallucinations might still happen, which could be especially problematic in biomedical use cases and might warrant additional manual fact checking \cite{nakov2021automated} before using the output of LLM-based systems in downstream tasks \cite{kim2024m}.

Another issue to consider is data privacy. These models might repeat their training data if prompted in a specific way \cite{nasr2023scalable}. That means training data has to be carefully anonymized before training or fine-tuning these models. The same problem arises for the few-shot examples, personal data should be removed from all context that these models might repeat. They also might just make up facts about people, which could put vendors and service providers at legal risk for defamation\footnote{\url{https://www.reuters.com/technology/australian-mayor-readies-worlds-first-defamation-lawsuit-over-chatgpt-content-2023-04-05/}}.

Another big ethical issue is job replacement and automation. Klarna, a financial service provider and one of the early enterprise customers of OpenAI published a report in February 2024 stating that their AI-powered customer support assistant handled one third of their customer support requests "doing the equivalent work of 700 full-time agents"\footnote{\url{https://web.archive.org/web/20240305093659/https://www.klarna.com/international/press/klarna-ai-assistant-handles-two-thirds-of-customer-service-chats-in-its-first-month/}}. This automation trend could not only lead to societal issues if more and more people are made redundant with LLM-powered systems, but also to quality issues when humans are taken out of the loop and more and more users and companies are trusting LLM-generated content without double-checking it\footnote{\url{https://web.archive.org/web/20240306115841/https://www.forbes.com/sites/mollybohannon/2023/06/08/lawyer-used-chatgpt-in-court-and-cited-fake-cases-a-judge-is-considering-sanctions/}}\footnote{\url{https://web.archive.org/web/20240304162744/https://www.bbc.com/travel/article/20240222-air-canada-chatbot-misinformation-what-travellers-should-know}}.

\section{Conclusion}\label{Conclusions}
We showed that a downloadable open-source model (Mixtral 8x7B Instruct V0.1) was competitive with some of the best available commercial models in a domain-specific biomedical RAG setting when used with 10-shot learning. This opens up the possibility to have state-of-the-art performance in use cases where using third-party APIs is not feasible because of the confidentiality of the data. The model used via a commercial hosting service was also significantly faster than Claude 3 Opus while being at least 30x cheaper.

We also observed that the zero-shot performance of this model was still lagging behind its commercial competitors, making it even unusable in some settings where highly specific structured output is required.

We were unable to achieve consistent performance improvements from QLoRa fine-tuning Mixtral or fine-tuning gpt-3.5-turbo via the proprietary fine-tuning service of OpenAI. This might be an indication that successfully fine-tuning these LLMs requires more engineering effort and costs and might not be worth the effort in some use cases compared to few-shot learning.

We tried to augment the context of these models with additional relevant knowledge from a knowledge base (Wikipedia), but again could not see consistent performance improvements. We speculate that this might be due to the knowledge in this setup being not novel enough for the models, or because of the way we combined it with few-shot learning. 

For future work we want to verify our results with different, more domain-specific tasks where less knowledge might have been present during the pre-training of these LLMs, and we also want to further explore optimal selection of few-shot examples, as this seems to get the best performance out of these models while being straightforward methodically.

\begin{acknowledgments}
 We want to thank the organizers of the BioASQ challenge for setting up this challenge and supporting us during our participation. We are also grateful for the feedback and recommendations of the anonymous reviewers.
\end{acknowledgments}

\bibliography{bibliography}

\appendix

\end{document}